\documentclass[conference]{IEEEtran}
\IEEEoverridecommandlockouts
\usepackage{url}

\usepackage{cite}
\usepackage{amsmath,amssymb,amsfonts}
\usepackage{graphicx}
\usepackage{textcomp}
\usepackage{algorithm}
\usepackage{algpseudocode}
\usepackage{graphicx}

\usepackage{xcolor}
\usepackage{lipsum,framed}
\def\BibTeX{{\rm 
B\kern-.05em{\sc 
i\kern-.025em b}
\kern-.08em
    T\kern-.1667em\lower.7ex\hbox{E}\kern-.125emX}}

\begin{document}

% \title{Option Batch-Shuffle Trick: Answering MCQs on Telecommunication Documents using Large Language Models*\\
% % \thanks{Identify applicable funding agency here. If none, delete this.}
% }

\title{AmaSQuAD: A Benchmark for Amharic Extractive Question Answering\\
}

\author{
\IEEEauthorblockN{Nebiyou Daniel Hailemariam, Blessed Guda, Tsegazeab Tefferi, }
\IEEEauthorblockA{\textit{College of Engineering, Carnegie Mellon University} \\
% Kigali, Rwanda (Blessed Guda, Gabrial Zencha A., Lawrence Francis) \\
% Pittsburgh, USA (Carlee Joe-Wong) \\
nhailema@alumni.cmu.edu, blessed@alumni.cmu.edu, ttefferi@alumni.cmu.edu
}

}

\maketitle
\begin{abstract}
This research presents a novel framework for translating extractive question-answering datasets into low-resource languages, as demonstrated by the creation of the AmaSQuAD dataset, a translation of SQuAD 2.0 into Amharic. The methodology addresses challenges related to misalignment between translated questions and answers, as well as the presence of multiple answer instances in the translated context. For this purpose, we used cosine similarity utilizing embeddings from a fine-tuned BERT-based model for Amharic and Longest Common Subsequence (LCS). Additionally, we fine-tune the XLM-R model on the AmaSQuAD synthetic dataset for Amharic Question-Answering. The results show an improvement in baseline performance, with the fine-tuned model achieving an increase in the F1 score from 36.55\% to 44.41\% and 50.01\% to 57.5\% on the AmaSQuAD development dataset. Moreover, the model demonstrates improvement on the human-curated AmQA dataset, increasing the F1 score from 67.80\% to 68.80\% and the exact match score from 52.50\% to 52.66\%.The AmaSQuAD dataset is publicly available Datasets\footnote{\url{https://huggingface.co/datasets/nebhailema/AmaSquad}}.
\end{abstract}

\section{Introduction}
The latest census results have shown that Amharic is spoken by approximately 57.5 million people in Ethiopia of which 25.1 million individuals within the country have adopted it as a second language \cite{hirpassa2023improving}. In addition, next to Arabic, Amharic is the second most widespread Semitic language \cite{hirpassa2023improving}. Despite being the second most widespread Semitic language, Amharic lacks substantial Natural Language Processing (NLP) resources and tools \cite{basha2023detection}. This shortage of resources has hindered the development of robust systems that could greatly benefit various natural language tasks.  

Question answering (QA) is a Natural Language Processing task that aims to accurately provide answers  to natural language questions \cite{zhu2021retrieving}. Unlike search engines, which present a list of relevant documents for a question, QA systems provide a definite answer, enhancing user experience. Search engines such as Google and Bing incorporate a QA system in their systems to respond precisely to questions \cite{zhu2021retrieving}.   
QA could be categorized into two tasks - Open-domain QA (OpenQA) and Machine Reading Comprehension (MRC) based on the contextual information given to the system. OpenQA is an approach that leverages large unstructured text data to provide answers to questions in natural language \cite{zhu2021retrieving}. The traditional question-answering pipeline in the OpenQA approach involves processes such as Document Retrieval and Answer Extraction with a component known as a Reader \cite{zhu2021retrieving}. During document retrieval, the goal is to identify relevant passages and documents that may contain answers to the question, employing techniques such as TF-IDF, BM25, or search engines. Following this, the answer is extracted from the identified and pertinent documents.  In the case of MRC, the aim is to read a document and extract the answer without having to do document retrieval. Similar to OpenQA, this is achieved by using the Reader component.  

The Reader component of a QA system can be either extractive or generative readers \cite{zhu2021retrieving}. Extractive readers extract answers by predicting a span from retrieved documents, while generative readers utilize sequence-to-sequence models to produce answers in natural language. Moreover, OpenQA systems have recently adopted modern architecture by incorporating neural Machine Reading Comprehension. A substantial amount of high-quality data has been generated to train OpenQA systems, including datasets like MS MARCO \cite{nguyen2016ms}, CNN/Daily Mail \cite{hermann2015teaching}, and SQuAD \cite{rajpurkar2018know}.

In contrast to the abundance of high-quality datasets available for widely spoken languages, the Amharic language faces a dataset scarcity. The largest Amharic dataset collected by \cite{abedissa2023amqa} includes only 2,628 sets of questions and answers . Consequently, to our knowledge, no expansive extractive QA dataset is currently available for the Amharic language. Without a sufficiently large and diverse dataset, it becomes challenging to train models capable of providing accurate responses, thus further impeding progress in the development of Amharic QA system development.

In response to the scarcity of datasets for training Amharic Question Answering Models, we build a translation-based data generation framework valuable for extractive QA. Recognizing the limited availability of comprehensive datasets in Amharic, the study employs Google Translate to transform the widely used SQuAD 2.0 training and development dataset into an Amharic dataset named AmaSQuAD. In addition to creating the AmaSQuAD dataset, we aim to leverage this dataset, which is specifically tailored for the extractive approach to train an MRC-based Amharic question-answering model. We aim to accomplish this by fine-tuning the XLM-R model. Furthermore, we anticipate that the developed framework can be used for other languages and other extractive QA datasets.

The rest of the research is structured into several sections. Section II provides a literature review of existing approaches for extractive QA, offering insights into previous research. Section III describes the methodology adopted in the research, detailing the SQuAD dataset description, the proposed translation framework, and the Extractive Question Answering Model. Following this, Section IV presents an analysis of the AmaSQuAD dataset and the XLM-R Fine-Tuning Process. Subsequently, Section 5 discusses the results of the comparative test using AmaSQuAD and AmQA. Finally, Section 6 offers a conclusion summarizing the key findings and suggesting directions for future research.

\section{Related Work}

\cite{abedissa2023amqa} introduced the Amharic Question Answering dataset, known as AmQA - the first publicly accessible dataset \cite{abedissa2023amqa}. They gathered 2,628 sets of questions and answers by leveraging crowdsourcing, drawing from a pool of 378 Wikipedia articles. Their primary objective was to facilitate extractive QA, providing the precise location within the context where answers to questions could be found. In an effort to establish a baseline for the AmQA dataset, the researchers employed an XLM-R model, fine-tuned on SQuAD 2.0—a cross-lingual Roberta model pre-trained on an extensive 2.5TB dataset spanning 100 languages \cite{conneau2019unsupervised}. They assessed two variations of the XLM-R-based QA model: XLM-R base and XLM-R large. Notably, XLM-R large outperformed XLM-R small, achieving an EM score of 50.76 and an F1 score of 71.74. 

In addition to evaluating the QA model in isolation, the researchers integrated a retriever-reader component to rank and filter relevant context for answering questions. This combined approach, featuring both the QA and retriever models, yielded an EM score of 50.3 and an F1 score of 69.58. One limitation of their work was that they did not conduct fine-tuning on any model using the AmQA dataset; instead, they solely relied on pre-existing models for inference and evaluation. Additionally, unlike SQuAD 2.0 the AmQA dataset doesn’t include unanswerable questions. Furthermore, specific details regarding the size of the evaluation set were not disclosed.

Lewis et al. aimed to address the challenge of insufficient relevant data for benchmarking multilingual models, in order to assess the performance of QA systems across various languages and promote progress in multilingual QA \cite{lewis2019mlqa}. The study introduced MLQA, a dataset that comprises QA pairs in seven languages. MLQA includes, on average, parallel QA instances across four languages. The study designed two tasks to evaluate the performance of QA models with MLQA, namely Cross-Lingual Transfer (XLT) and Generalized Cross-Lingual Transfer (G-XLT). In the case of XLT, the QA model is trained using context, query, and answer pairs from one language and then tested on context, query, and answer pairs from another language. On the other hand, in the case of G-XLT, the model is similarly trained, but its performance is evaluated based on its ability to answer questions by extracting answers from a context in a different language. In G-XLT, it's important to note that the extracted answer is not in the same language as the query. The assumption in this evaluation is that the parallel dataset can enable comparison across different languages. The study evaluated state-of-the-art cross-lingual models and machine translation-based models on MLQA and set a baseline. 

To tackle the challenges of training a QA model with limited data, Carrino et al. devised the Translate Align Retrieve (TAR) method. This technique involves automatically translating the SQuAD 1.1 dataset into Spanish to generate synthetic corpora \cite{carrino2019automatic}. The researchers trained a Neural Machine Translation model from scratch to translate the context, query, and answers from the English SQuAD dataset into Spanish. They then utilized EFMARAL to calculate the alignment between the source translation and the context, facilitating answer retrieval from the translated text \cite{ostling2023efmaral}. The success of answer retrieval in the translated context is contingent on the alignment between the source and translated contexts. After implementing this approach to translate the SQuAD 1.1 dataset, the authors fine-tuned a pre-trained multilingual BERT (mBERT) model. The study evaluated the performance using two corpora, namely MLQA and XQuAD. The TAR-based training approach demonstrated superior F1 scores compared to XLM on the MLQA dataset and set a new state-of-the-art standard on the XQuAD dataset. This evaluation affirmed that TAR-based mBERT outperformed mBERT-based models validating the significance of the proposed approach.

\cite{abedissa2023amqa} developed a non-factoid, rule-based QA system in Amharic, tailored to address diverse information needs, including biography, definition, and description questions \cite{abedissa2019amharic}. This text-based QA strategy employs a hybrid approach for question classification, utilizing word overlap with the query for document filtering and rule-based scoring for answer extraction. Specifically, a summarization technique is applied for biography questions, and the resulting summary is validated using a text classifier. The performance of the answer extraction component yielded an F1 score of 0.592%. Notably, the dataset used for training and evaluating the system has not been publicly released. Moreover, the study did not conduct a comparative analysis of the rule-based QA extraction approach with other approaches.

\cite{abadani2021parsquad} presented ParSQuAD, a QA dataset derived from the SQuAD 2.0 dataset using machine translation \cite{abadani2021parsquad}. The study used Google Translate to translate the dataset into Persian. The authors faced two challenges after translating the dataset: finding the starting index of the translated answer in the translation context and finding translations that contain errors in either the context, query, or answer, which is more challenging. The authors created two datasets after the translation. One that has been manually edited and another generated automatically. In both cases, when locating the answer span in the translated context, the researchers used the one-to-one mapping between the source and the translated sentence to locate the answer in the translated sentence. Subsequently, the researchers fine-tuned three BERT-based pre-trained models: ParsBERT, ALBERT, and mBERT. The models were fine-tuned using default parameters from the Huggingface script and trained for two epochs. The study evaluates the three models on three variations of the ParSQuAD dataset: automatically generated, manually edited, and a variation of the manually edited dataset with extended unanswerable questions. The authors were able to set a baseline model, with the mBERT model outperforming the other two.

\cite{clark2019boolq} introduced BoolQ - a dataset containing passages, and questions with corresponding Yes or No answers \cite{clark2019boolq}]. Answering these questions necessitates an understanding of what an assertion in a passage entails, including what it excludes and makes improbable. Such questions are often posed when individuals are seeking more detailed information. Providing accurate responses to Boolean questions requires a wide range of inferential abilities.  In their research, various methods were explored to construct a polar question-answering model for the BoolQ dataset. A majority baseline, which consistently predicted the majority class, yielded an accuracy of 62\%. A recurrent model, without any pretraining, achieved an accuracy of 69.6\%. Subsequently, the researchers used transfer learning by incorporating datasets such as QNLI, SQuAD 2.0, NQ Long Answer, and MultiNLI. They applied fine-tuning and conducted experiments with pre-trained models, including the recurrent model, OpenAI GPT, and BERT. Among the various models, the unsupervised BERT model, with multi-step pretraining on Books and Wikipedia, followed by additional pretraining on MultiNLI data, and subsequent fine-tuning on the BoolQ dataset, demonstrated the highest performance, achieving an accuracy of 80.43\%. There remains a significant gap between human annotators, who have an accuracy of 90\%. 

The literature highlights a notable data scarcity in training models for low-resource languages like Amharic. To address this challenge, a viable approach involves leveraging neural machine translation as \cite{abadani2021parsquad} and Carrino et al. and implementing post-processing techniques \cite{carrino2019automatic}, \cite{abadani2021parsquad}. This enables the creation of data derived from resource-rich languages, such as English, which can then be used for training a question-answering model. This strategy leverages the abundance of resources available in a well-supported language such as English to enhance the development of question-answer models for languages with limited available data like Amharic.

\section{Proposed Methodology}
\subsection{SQuAD Dataset Description}
Squad 2.0 is one of the widely used datasets in QA. Unlike its predecessor, SQuAD 1.1, SQuAD 2.0 introduces the concept of unanswerable questions, requiring models to not only understand the context but also identify when a question lacks a definite answer in a given passage [6]. The dataset has the following properties: id, context, question, and answers \cite{squad_v2_huggingface}. The "id" field serves as a unique identifier for the question. The "context" field contains a passage of text from Wikipedia from which questions serve as the context for deriving the question and answer. The "question" field is where the specific query related to the given context is presented. The "answers" field consists of information about the answer to the question, with "answer\_start" indicating the character position in the context where the answer begins and "text" containing the actual answer text.

\subsection{Proposed Translation Framework}
The proposed methodology for translating the SQuAD 2.0 dataset into AmSQuAd involves several key steps. For each entry in the SQuAD 2.0 dataset, we extract information such as the title, context, question, and answers. We then proceed to translate the title, context, and question into Amharic using Google Translate through the Deep Translator Python library. However, two main issues need to be addressed during this process.  

The first issue relates to the translated answers not always aligning perfectly with the content in the translated context. In other words, the translated answer may not be located in the translated context verbatim. This misalignment between the question and the answer in the translated context can pose a significant challenge.  The second challenge relates to the possibility of the translated context containing multiple instances of the answer. This can lead to ambiguity when determining the correct starting index for the answer within the context. To mitigate this issue, a systematic approach is required to identify the most relevant instance of the answer and accurately mark its starting position in the translated context.  

To address the first issue, we employ cosine similarity between the embeddings of the text span in the translated context and the embeddings of the translated answer to determine their similarity. The embeddings are derived from fine-tuning a BERT-based model (specifically, bert-base-multilingual-cased) using Amharic data \cite{davlan_bert_amharic}. But this alone is not sufficient in identifying if the span in the translated context is similar to the translated answer. To further measure the similarity between the two, we do so by first identifying the length of the Longest Common Subsequence (LCS) shared between the span and the answer sequences \cite{cormen2022introduction}. Subsequently, we calculate the similarity percentage by dividing this common subsequence length by the maximum length of the two input sequences. This percentage reflects how much the two sequences overlap or have in common, providing a measure of their similarity.  The similarity scores are used to select the answer span in the context.

To address the second challenge of dealing with translated contexts that may contain multiple instances of the answer, the approach aligns closely with the methodology employed in the research conducted by N. Abadani et. al. In their study, the authors devised a method for extracting the translated answer from the translated context \cite{abadani2021parsquad}. This involved a two-step process: first, they identified the sentence within the original context where the original answer was originally located. Next, they attempt to extract the starting index of the translated answer from the translated context.   

Similarly, in this research, when extracting the answer, we first locate the sentence in which the answer was originally found in the original context. We then select spans in the translated context that are in close proximity to the starting index of the answer in the original text. This is based on our intuition that the answer ordering in the translated Amharic context won't deviate significantly from the answer in the original English context. Therefore, translated answer spans that are in close proximity to the sentence where the answer was originally located in the English context are prioritized.  

As shown in Algorithm 1., the entire process followed in this research to extract the translated answer from the translated context applies a weighted combination of two operations: cosine similarity and the Longest Common Subsequence. It scans the entire context while prioritizing the translated answers that are in close proximity to the original answer. The weights assigned to the two operations for the most similar span are w1 and w2, respectively, with values of 2/3 and 1/3.

\begin{algorithm}[H]
\small
\caption{Proximity, Similarity, and LCS Amharic Answer Extractor}
\textbf{Input:} \( SQuAD\ 2.0 \) entries \( (to, co, qo, ao) \) \\ 
\( (title_{orig}, context_{orig}, query_{orig}, answer_{orig}) \)\\
\textbf{Output:} Translated entries \( (tt, ct, qt, at) \)\\
\( (title_{tr}, context_{tr}, answer_{tr}) \)

\begin{algorithmic}[1]
\State \textbf{Define embedding and LCS weights:} \( w_1 = \frac{2}{3}, w_2 = \frac{1}{3} \)

\For{\textbf{each} \( (to, co, qo, ao) \)}
    \State \textbf{Translate fields:}
    \State \( tt \gets tr(to), \quad ct \gets tr(co) \)
    \State \( qt \gets tr(qo), \quad at \gets tr(ao) \)

    \Statex \textbf{Initialize variables:}
    \State \( s_{max} \gets 0, \quad idx_b \gets -1 \)
    \State \( sent_b \gets "", \quad p_b \gets \infty \)

    \Statex \textbf{Split context and answer into words:}
    \State \( c_w \gets sp(ct), \quad a_w \gets sp(at) \)
    \State \( a_emb \gets emb(at) \)

    \Statex \textbf{Scan context for best match:}
    \For{\( i = 0 \) \textbf{to} \( \text{len}(c_w) - \text{len}(a_w) \)}
        \For{\( s = 0 \) \textbf{to} \( 3 \)} \Comment{Stride over context}
            \State \( w_txt \gets ext_w(c_w, i, \text{len}(a_w), s) \)
            \State \( w_emb \gets emb(w_txt) \)
            \State \( sim \gets w_1 \cdot sim(a_emb, w_emb) + w_2 \cdot lcs(w_txt, at) \)

            \If{\( sim > s_{max} \)}
                \If{\( prox(at, w_txt) < p_b \)}
                    \State \( s_{max} \gets sim, \quad idx_b \gets st(c_w, i) \)
                    \State \( sent_b \gets w_txt, \quad p_b \gets prox(at, w_txt) \)
                \EndIf
            \EndIf
        \EndFor
    \EndFor

    \Statex \textbf{Store best match:}
    \State \( at \gets sent_b, \quad ats \gets idx_b \)
    \State Save \( (tt, ct, qt, at, ats) \)
\EndFor
\end{algorithmic}
\end{algorithm}

\subsection{Extractive Question Answering Models}

There have been significant improvements in language models designed for low-resource languages. Models such as mBERT, XLM, and XLM-RoBERTa (XLM-R) have played a pivotal role in enhancing capabilities, not only for languages with abundant resources but also for those facing resource limitations  \cite{devlin-etal-2019-bert}, \cite{liu2019roberta}, \cite{pires-etal-2019-multilingual}.\cite{pires-etal-2019-multilingual}, illustrated that a BERT model can be effectively pre-trained on 104 languages using Wikipedia with good performance in both few-shot and zero-shot model evaluations \cite{liu2019roberta}. However, it's worth noting that the model was not pre-trained in the Amharic language.   

XLM-R, a transformer-based masked language model trained using 2.5 terabytes of CommonCrawl data, made significant improvements across various tasks\cite{conneau2019unsupervised}. With a substantial vocabulary size of approximately 250 thousand, XLM-R two model variants: XLM-R Base with 270 million parameters and XLM-R Large with 550 million parameters. Its performance on the MLQA dataset, evaluated across seven languages, deserves attention. It achieves an average F1 score and exact match of 70.7\% and 52.7\%, respectively, surpassing XLM-15, which scored 61.6\% on the F1 score and 43.5\% on the exact match. Additionally, XLM-R outperforms mBERT by a considerable margin of 13.0\% on the F1-score and 11.1\% on the exact match matrix. Furthermore, in the monolingual comparison, XLM-R has demonstrated superior performance compared to BERT Large trained on monolingual data. It shows the robust capacity of the model.  

It's important to note that XLM-R is also pre-trained on Amharic language. For this reason, we further leverage XLM-R Large by fine-tuning the model specifically for Amharic QA tasks. Leveraging XLM-R fine-tuned on SQuAD 2.0, we aim to improve the performance of the Amharic QA model.

\section{Experiment Setup}
The fine-tuning process of XLM-R on the Amharic Question-Answering (QA) task involves a three-epoch training cycle using the "xlm-roberta-large-squad2" base language model. The maximum sequence length is set to 256. A learning rate of 1e-5, while using the Cosine Annealing learning rate scheduler. The fine-tuning leverages the AmaSQuAD synthetic dataset to enhance the model's performance on the Amharic QA task.

We trained the XLM-R model using the AmaSQuAD dataset. Refer to Section 5.1 for a more detailed analysis of the translated data sets. We first filtered the dataset. The selection process for training data involves evaluating the answers associated with Amharic SQuAD 2.0 questions. We consider the answer for inclusion based on a similarity criterion. Specifically, answers with a similarity probability of 0.6 or higher are included. This filtering ensures that the train and test datasets are of higher quality.

The filtered training dataset contains about 57,650 answerable questions, while the development data set contains about about 13,779. Moreover, AmaSquad contains over 50,000 unanswerable questions, which do not have any similarity score, we downsampled them to 6000 for the training set and 700 for the development set. In total, the training set contained 63,650 instances, while the development set contained 14,479 instances.

\section{Results and Discussion}

\subsection{	Analysis of Translated Answers in AmaSQuAD}

\begin{figure}[h!]
    \centering
    \includegraphics[width=1.0\linewidth]{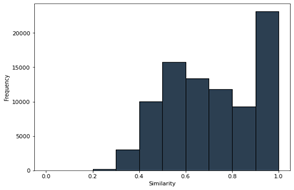}
    \caption{Distribution of Similarity on a scale from 0 to 1 of translated answers with the translated context span for the AmaSQuAD training set}
    \label{fig:training}
\end{figure}

\begin{figure}[h!]
    \centering
    \includegraphics[width=1.0\linewidth]{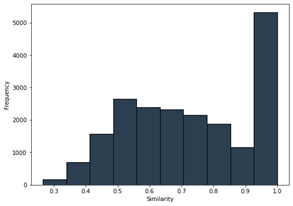}
    \caption{Distribution of Similarity on a scale from 0 to 1 of translated answers with the translated context span for the AmaSQuAD development set}
    \label{fig:development}
\end{figure}

Figures \ref{fig:training} and \ref{fig:development} show the similarity distribution results on a scale from 0 to 1 of translated answers with the translated context span for AmaSQuAD. The frequency sharply drops in the range of 0 to 0.4, implying a rare occurrence of translated answers within minimal similarity in this range. The frequency distribution indicates that a substantial number of translations fall within the similarity range of 0.4 to 1.0, peaking at 0.9 to 1.0. As the similarity increases, the frequencies mostly rise, peaking at 23,144 and 5314 occurrences in the range of 0.9 to 1.0 for the training and development dataset, respectively. This indicates more accurate translations closely aligned with the original context in the translated context that have high similarity ranges.

Various adjustments of weights \( w_1 \) and \( w_2 \) for cosine similarity and LCS should be considered for Algorithm 1. Tuning these parameters involves a tradeoff between focusing on the semantic and syntactic similarities between translated answers and their corresponding context spans. Additionally, incorporating human feedback on the quality of the translated AmaSQuAD dataset is essential. Human evaluations can provide valuable insights into areas where the algorithm may falter or produce inaccurate results, allowing for targeted improvements to be made.

\begin{table*}[h!]
\centering
\begin{tabular}{|l|c|c|c|}
\hline
\textbf{Aspect}           & \textbf{Before Fine-Tuning} & \textbf{After Fine-Tuning} & \textbf{Performance Increase} \\ \hline
\textbf{Dataset}          & AmaSQuAD (Synthetic)       & AmaSQuAD (Synthetic)       &                                \\ \hline
\textbf{Exact Match (EM)} & 36.55\%                    & 44.41\%                    & \textbf{+7.86\%}              \\ \hline
\textbf{F1-Score}         & 50.01\%                    & 57.55\%                    & \textbf{+7.54\%}              \\ \hline
\textbf{Dataset}          & AmQA (Human-Curated)       & AmQA (Human-Curated)       &                                \\ \hline
\textbf{Exact Match (EM)} & 52.50\%                    & 52.66\%                    & \textbf{+0.16\%}              \\ \hline
\textbf{F1-Score}         & 67.80\%                    & 68.80\%                    & \textbf{+1.00\%}              \\ \hline
\end{tabular}
\vspace{1mm}  % Adds 5mm of space between the table and the caption
\caption{Performance Comparison of XLM-R Before and After Fine-Tuning}
\label{tab:performance_comparison}
\end{table*}

\subsection{Performance Analysis of XLM-R on AmaSQuAD and AmQA}
The evaluation presented in Table 1 demonstrates the improvement in the performance of the fine-tuned XLM-R model on both synthetic and human-curated datasets, namely AmaSQuAD and AmQA, respectively. Initially, without fine-tuning, the model achieved low scores on both datasets. However, after fine-tuning on AmaSQuAD synthetic data, there was a notable increase in performance. Specifically, the Exact Match and F1-score increased from 36.55 to 44.41 percent and from 50.01 to 57.55 percent, respectively, on the AmaSQuAD development data. Similarly, on the AmQA dataset, the Exact Match improved from 52.50 to 52.66 percent, and the F1 score increased from 67.80 to 68.80 percent post-training. This suggests that leveraging synthetic data for fine-tuning can indeed augment the model's ability to reason and answer questions effectively without compromising performance, as evidenced by the slight increase in the F1-score on the human-curated AmQA dataset.

Unlike the baseline model set by \cite{abedissa2019amharic}, which lacks clarity regarding the percentage of training, validation, and test data used for fine-tuning and evaluation of the XLM-R model, this study is the first to establish a detailed baseline performance using the AmQA dataset, which contributes to the literature on question answering in underrepresented languages. The hypothesis that fine-tuning with synthetic data would improve the model's performance on both synthetic and human-curated datasets is partially proven. While there was an improvement in performance on both datasets, the extent of improvement was relatively small on the AmQA data. Additionally, it's essential to acknowledge the weaknesses of the AmQA dataset, such as the absence of unanswerable questions, which limits its representativeness of real-world scenarios.

The absence of unanswerable questions in the AmQA dataset is a notable limitation, as real-world scenarios often involve questions for which no explicit answer exists in the provided context, which means that the performance assessment done on the AmQA dataset only assessed the performance of the model on answerable questions. Moreover, the dataset size of 2,628 is relatively small, which could impact the generalizability of the findings. 

Therefore, future model evaluation should incorporate human-curated datasets with a more diverse range of question types to provide a more comprehensive assessment of the model's capabilities in real-world settings. Additionally, to further advance the state-of-the-art in question answering for underrepresented languages like Amharic, it is imperative to explore the impact of pre-training on a large Amharic corpus with various pre-training strategies, as presented in \cite{liu2019roberta}. Incorporating such pre-training techniques has been shown to enhance a model's understanding and performance on downstream tasks like question answering. Addressing these limitations would be crucial for future research to further advance the state-of-the-art in question answering.

\section{Conclusion}
In conclusion, we present a comprehensive framework for translating the SQuAD 2.0 dataset into Amharic, resulting in the creation of the AmaSQuAD dataset. The proposed methodology leverages Google Translate through the Deep Translator Python library for language translation, addressing the challenges of misalignment between translated questions and answers, as well as the presence of multiple instances of answers in the translated context.

To enhance the accuracy of the translation, we utilize techniques such as cosine similarity based on embeddings from a BERT-based model fine-tuned on Amharic and Longest Common Subsequence (LCS). The approach prioritizes answer spans near the original answer's location in the English context, mitigating potential discrepancies in locating translated answers in translated contexts.

The research fine-tunes the XLM-R model on the AmaSQuAD synthetic dataset for Amharic Question-Answering tasks. This study is the first to establish a baseline performance using the AmQA dataset, The fine-tuned model improved the baseline performance on both the AmaSQuAD development dataset from  36.55% to 44.41% and  50.01% to 57.55% on the f1-score and exact match score, respectively. On the human-curated AmQA dataset, it improved the performance from 67.80% to 68.80% and 52.50% to 52.66% on the f1-score and exact match score, respectively. This shows the effectiveness of leveraging synthetic data in enhancing the model's ability to answer questions that it has not encountered before while also gaining a 1% increase in the F1 score on the human-curated AmQA dataset.

Overall, the proposed translation framework, coupled with the fine-tuned XLM-R model, not only contributes to the availability of a valuable resource in the form of the AmaSQuAD dataset but also demonstrates the potential for advancing the capabilities of Amharic Question-Answering models using synthetic data

\bibliographystyle{IEEEtran}
\bibliography{references}

\end{document}